\pdfoutput=1

\documentclass[11pt]{article}

\usepackage[]{acl}

\usepackage{times}
\usepackage{latexsym}

\usepackage[T1]{fontenc}

\usepackage[utf8]{inputenc}

\usepackage{microtype}
\usepackage{algorithm}
\usepackage{algorithmic}
\usepackage{times}
\usepackage{helvet}
\usepackage{courier}
\usepackage{microtype}
\usepackage{amsfonts}
\usepackage{amssymb}
\usepackage{multirow}
\usepackage{graphicx}
\usepackage{amssymb}
\usepackage{natbib}
\usepackage{amsmath}
\usepackage{appendix}
\usepackage{subfigure}
\usepackage{booktabs}
\usepackage{color}
\usepackage{pifont}
\usepackage{subfigure}
\usepackage{graphicx}
\usepackage{makecell}
\usepackage{color, colortbl}
\title{Generalized Category Discovery with Large Language Models in the Loop}

\author{
Wenbin An$^{1,3}$, Wenkai Shi$^{1,4}$, Feng Tian$^{2,3 *}$, Haonan Lin$^{2,4}$, QianYing Wang$^{5 *}$ \\ \textbf{Yaqiang Wu$^{5}$, Mingxiang Cai$^{5}$, Luyan Wang$^{5}$, Yan Chen$^{2,4}$, Haiping Zhu$^{2,4}$, Ping Chen$^{6}$} \\
$^1$ School of Automation Science and Engineering, Xi'an Jiaotong University \\
$^2$ School of Computer Science and Technology, Xi’an Jiaotong University \\
$^3$ Ministry of Education Key Laboratory of Intelligent Networks and Network Security \\
$^4$ Shaanxi Province Key Laboratory of Big Data Knowledge Engineering \\
$^5$ Lenovo Research
$^6$ University of Massachusetts Boston \\
\texttt{wenbinan@stu.xjtu.edu.cn,fengtian@mail.xjtu.edu.cn} \\ 
}

\begin{document}
\maketitle
\begin{abstract}
\renewcommand{\thefootnote}{\fnsymbol{footnote}}
\footnotetext[1]{Corresponding Authors.}
\renewcommand{\thefootnote}{\arabic{footnote}}
Generalized Category Discovery (GCD) is a crucial task that aims to recognize both known and novel categories from a set of unlabeled data by utilizing a few labeled data with only known categories. Due to the lack of supervision and category information, current methods usually perform poorly on novel categories and struggle to reveal semantic meanings of the discovered clusters, which limits their applications in the real world. To mitigate the above issues, we propose \textit{Loop}, an end-to-end active-learning framework that introduces \textit{Large Language Models (LLMs)} \footnote{The LLMs can be either locally deployed models or LLM APIs. In this paper, we use OpenAI's APIs for simplicity.} into the training loop, which can boost model performance and generate category names without relying on any human efforts. Specifically, we first propose \textit{Local Inconsistent Sampling (LIS)} to select samples that have a higher probability of falling to wrong clusters, based on neighborhood prediction consistency and entropy of cluster assignment probabilities. Then we propose a \textit{Scalable Query} strategy to allow LLMs to choose true neighbors of the selected samples from multiple candidate samples. Based on the feedback from LLMs, we perform \textit{Refined Neighborhood Contrastive Learning (RNCL)} to pull samples and their neighbors closer to learn clustering-friendly representations. Finally, we select representative samples from clusters corresponding to novel categories to allow LLMs to generate category names for them. Extensive experiments on three benchmark datasets show that \textit{Loop} outperforms SOTA models by a large margin and generates accurate category names for the discovered clusters. Code and data are available at \url{https://github.com/Lackel/LOOP}.
\end{abstract}

\begin{figure}
\centering
\includegraphics[width=6cm]{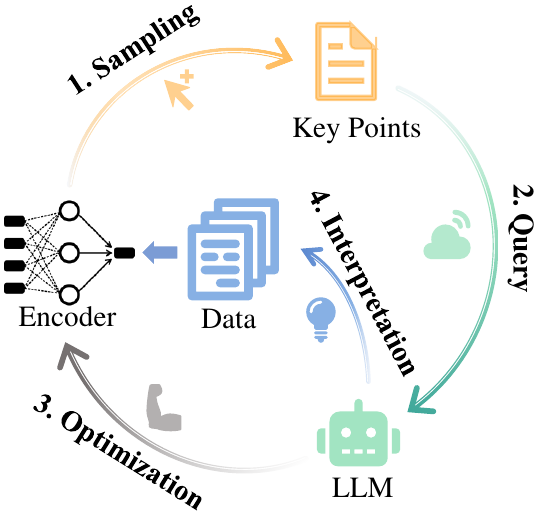}
\caption{The training loop of our model.} 
\label{fig1}
\end{figure}

\section{Introduction}
Although modern machine learning systems have achieved superior performance on many tasks, the vast majority of them follow the closed-world setting that assumes training and test data are from the same set of pre-defined categories \citep{openworld}. However, in the real world, many practical problems such as intent detection \citep{ptjn,shi2023diffusion} and image recognition \citep{ncl,semantic} are open-world, where the well-trained models may encounter data with unseen novel categories. To cope with this limitation, Generalized Category Discovery (GCD) was proposed and widely studied in both NLP \citep{thu2021,dpn} and computer vision \citep{gcd,simple}.
GCD requires models to recognize both known and novel categories from a set of unlabeled data, based on some labeled data containing only known categories, which can adapt models to the emerging categories without any human efforts.

Current methods \citep{dpn,simple,gcd,tan,ktn} usually first perform supervised pretraining on labeled data and self-supervised learning on unlabeled data to train a base model such as BERT \citep{bert}, then they perform clustering methods such as KMeans to discover both known and novel categories. Even though these methods can improve performance on known categories, they usually perform poorly on novel categories due to the lack of supervision. Furthermore, they also struggle to reveal semantic meanings (e.g., category names or descriptions) of the discovered clusters due to the lack of prior knowledge for novel categories. Recently, Large Language Models (LLMs) such as ChatGPT have shown extraordinary capabilities for various applications even without any labeled samples \citep{chatzero}. However, LLMs cannot be directly applied to GCD which requires models to cluster thousands of samples. And problems such as data privacy, high inference latency, and high API cost also limit their applications in the real world. 

\begin{figure*}
\centering
\includegraphics[width=14.5cm]{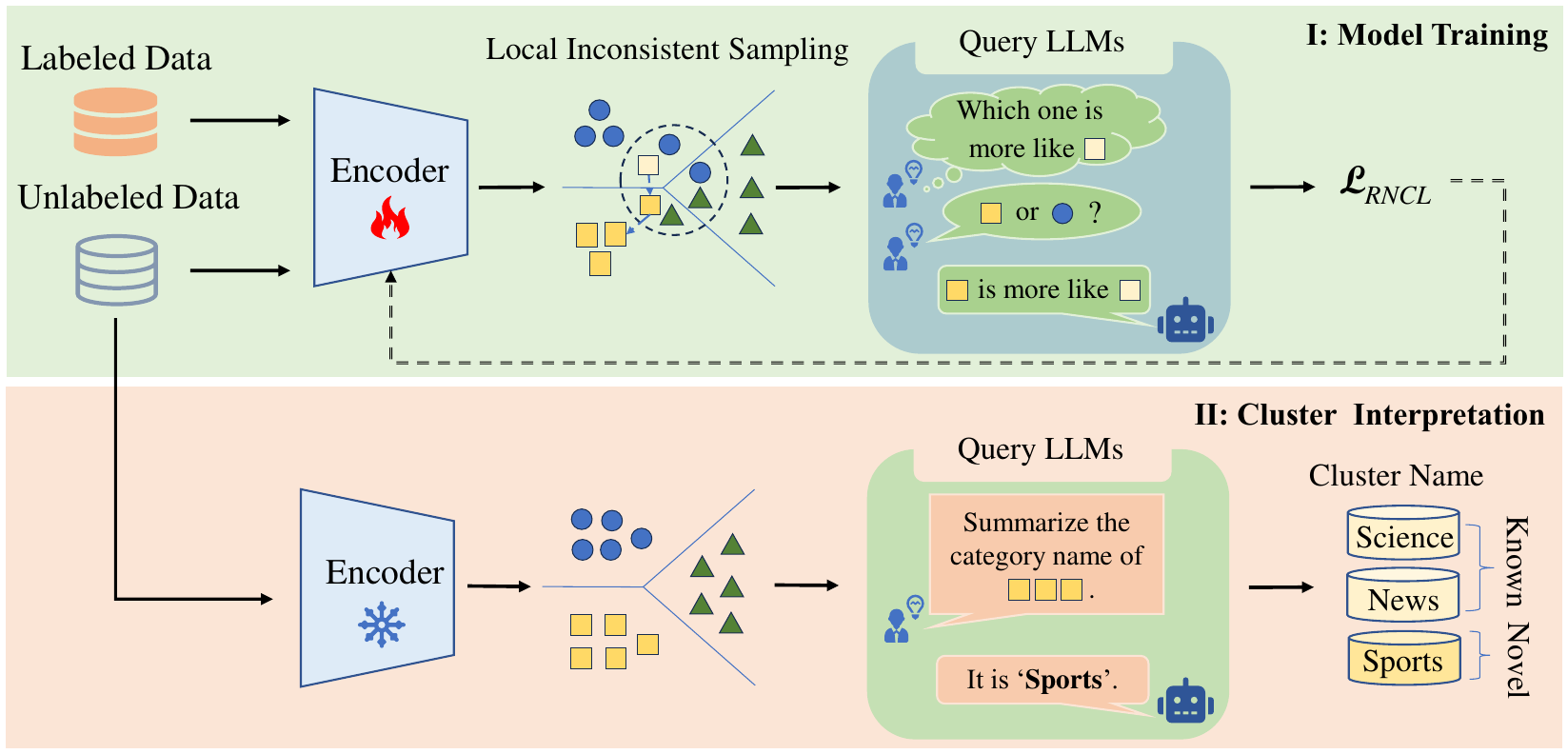}
\caption{The overall architecture of our model.} 
\label{fig2}
\end{figure*}

To solve the above limitations and enjoy the benefits of both base models and LLMs, we propose \textit{Loop}, an end-to-end active-learning framework that introduces LLMs into the training process. By selecting a few key samples to query LLMs and optimizing the base model based on the feedback, \textit{Loop} can compensate for the lack of supervision and generate category names for the discovered clusters with little query cost. Hence, we only need to train and maintain a small base model locally, which can reduce the inference cost and protect data privacy.
Specifically, as shown in Fig. \ref{fig1}, we first propose \textit{Local Inconsistent Sampling (LIS)} to select the most informative samples that have a higher probability of falling into wrong clusters. Specifically, we select samples that have high entropy of cluster assignment probabilities and whose neighbors have the most diverse cluster assignments. Intuitively, samples with high entropy and diverse neighbor predictions seem to violate the clustering assumption \citep{assumption} and locate near decision boundaries (Fig. \ref{fig2} dashed circle), so these neighbor-chaotic samples with great uncertainty would have a high probability of falling to wrong clusters \citep{mhpl}, so correcting them can significantly improve the model performance. 

After selecting the key samples, we need to build proper prompts to query the LLMs. However, due to the lack of information for novel categories, we cannot directly query LLMs which category these samples belong to as in traditional active learning. To solve this issue, we propose a \textit{Scalable Query} strategy that allows LLMs to choose true neighbors of the selected samples from multiple candidate neighbor samples \citep{clusterllm}. Based on the feedback of LLMs, we can solve the local inconsistency problem and decide which clusters these key samples truly belong to. Furthermore, LLMs are more competent at comparing semantic similarities between sentences than choosing from multiple category names. Then based on the refined neighbor relationships, we perform \textit{Refined Neighborhood Contrastive Learning (RNCL)} to pull samples closer to their neighbors to learn clustering-friendly representations. In this way, we can correct these samples by pulling them closer to the true clusters they belong to and cluster the rest of the samples to form more compact clusters. Finally, we decouple the clusters corresponding to novel categories from the discovered clusters \citep{dpn} and select a few samples closest to each center of the clusters to query LLMs to generate category names for novel categories.

Experimental results on three benchmark datasets show that \textit{Loop} outperforms SOTA models by a large margin and generates accurate category names for the discovered clusters. Furthermore, we also validate that the proposed \textit{Local Inconsistent Sampling} strategy can select more informative samples and the proposed \textit{Scalable Query} strategy can help to correct the selected samples effectively with minimal query cost.

Our contributions can be summarized as follows:
\begin{itemize}
  \item \textbf{Perspective}: we propose to introduce LLMs into the training loop to enjoy the benefits of both base models and LLMs. To the best of our knowledge, we are the first to utilize LLMs to guide the training process of GCD.

  \item \textbf{Framework}: we propose \textit{Loop}, an end-to-end active-learning framework that can select informative samples with \textit{Local Inconsistent Sampling} and label these samples with \textit{Scalable Query} without any human efforts.
  
  \item \textbf{Interpretation}: \textit{Loop} can reveal semantic meanings of the discovered clusters by generating category names for them, which is infeasible in previous methods.
  
  \item \textbf{Experiments}: Extensive experiments on three benchmark datasets show that \textit{Loop} outperforms SOTA models by a large margin (average 7.67\% improvement) and generates accurate category names with minimal query cost (average \$0.4 for each dataset).
\end{itemize}
  
\section{Related Work}
\subsection{Generalized Category Discovery}
Under the open-world assumption, GCD \citep{gcd} requires models trained on a few labeled data with known categories to recognize both known and novel categories, or even more fine-grained categories \citep{fcdc,dna,down} from the newly collected unlabeled data. Previous methods mainly performed representation learning on unlabeled data with self-supervised learning \citep{gcd, ncl} or pseudo-label learning \citep{dtc, mutual}. For example, \citet{thu2020} and \citet{ptjn} proposed to generate pseudo labels by clustering. \citet{mtp} performed contrastive learning to learn clustering-friendly representations. \citet{dpn} proposed to decouple known and novel categories with a prototypical network. \citet{ktn} and \citet{tan} proposed to transfer knowledge from known to novel categories to mitigate category bias caused by pretraining.

\subsection{Active Learning}
Active Learning (AL) aims to select informative samples for manual labeling to balance model performance and annotation cost. Previous methods are mainly based on prediction uncertainty (e.g., entropy \citep{clusterllm}, confidence \citep{confidence} and margin \citep{margin}) or sample information (e.g., MHPL \citep{mhpl} and CAL \citep{cal}). Recently, \citet{clusterllm} and \citet{contrastive} also utilized LLMs to replace human experts to save the annotation cost. However, how to employ active learning with LLMs in the open-world setting has not yet been explored.

\section{Method}
\noindent \textbf{Problem Setup.}\quad
Under the open-world assumption, models trained on a labeled dataset $\mathcal{D}^{l} = \{(x_{i},y_{i})|y_{i} \in \mathcal{Y}_{k}\}$ containing only known categories $\mathcal{Y}_{k}$ may encounter newly collected unlabeled data $\mathcal{D}^{u} = \{x_{i}|y_{i} \in \{\mathcal{Y}_{k}, \mathcal{Y}_{n} \}\}$ with both known categories $\mathcal{Y}_{k}$ and novel categories $\mathcal{Y}_{n}$, which can make the models fail. To cope with this challenge, Generalized Category Discovery (GCD) requires the models to recognize both known and novel categories based on $\mathcal{D}^{all} = \mathcal{D}^{l} \cup \mathcal{D}^{u}$, without any annotation or category information for novel categories. 
Finally, model performance will be measured on a testing set $\mathcal{D}^{t} = \{(x_{i},y_{i})|y_{i} \in \{\mathcal{Y}_{k}, \mathcal{Y}_{n} \}\}$.

\noindent \textbf{Framework Overview.}\quad
As shown in Fig. \ref{fig2}, there are two stages in the proposed \textit{Loop} framework. In the first stage, we introduce LLMs to guide the base model to learn better representations. Specifically, we first pre-train the base model for warm-up (Sec. \ref{pre}). Then we select informative samples for annotation based on \textit{Local Inconsistent Sampling} (Sec. \ref{sample}). Next, we construct the query prompt with the \textit{Scalable Query} strategy to query LLMs and acquire correct neighborhood relationships between samples (Sec. \ref{query}). Finally, we perform \textit{Refined Neighborhood Contrastive Learning} to learn clustering-friendly representations based on the feedback of LLMs (Sec. \ref{rncl}). In the second stage, we interpret the discovered clusters by decoupling and generating category names for the novel categories (Sec. \ref{interpret}).

\subsection{Multi-task Pre-training}
\label{pre}
We use the lightweight language model BERT \citep{bert} as the base model to extract features $z_{i} = F_{\theta}(x_{i})$ for the input sentence $x_{i}$. To quickly adapt the base model to current tasks, we pre-train $F_{\theta}$ on both labeled and unlabeled data in a multi-task manner \citep{mtp} with Cross-Entropy (CE) loss and Masked Language Modeling (MLM) loss \citep{bert}:
\begin{equation}
    \mathcal{L}_{pre} = \mathcal{L}_{ce}(\mathcal{D}^{l}) + \mathcal{L}_{mlm}(\mathcal{D}^{all})
\end{equation}
Through pretraining, $F_{\theta}$ can acquire both category-specific knowledge and general knowledge from the data, which can provide a good representation initialization for subsequent training.

\subsection{Local Inconsistent Sampling}
\label{sample}
To select informative samples that have a higher probability of falling to wrong clusters, we propose \textit{Local Inconsistent Sampling (LIS)} to select samples that make different predictions from their neighbors and have high prediction entropy. 

Specifically, we first perform Kmeans clustering on $\mathcal{D}^{all}$ to calculate cluster centers $\{\mu_{i}\}_{i=1}^{K}$ and get pseudo labels $\{\hat{y}_{j}\}_{j=1}^{N}$ for all data based on cluster assignments, where $K = |\mathcal{Y}_{k}| + |\mathcal{Y}_{n}|$ is the number of categories and $N=|\mathcal{D}^{all}|$ is the number of samples. We assume $K$ is known for a fair comparison with previous models and estimate it in Sec. \ref{estimate}.
Then for each feature $z_{i}$, we search its $k$-nearest neighbors in the feature space and denote $\mathcal{N}_{i}$ as the index set of the retrieved neighbors:
\begin{equation}
    \mathcal{N}_{i} = \mathop{argtop_{k}}_{j}\{sim(z_{i}, z_{j}) | j=1,...,N \}
\end{equation}
where $sim()$ is the cosine similarity function $sim(z_{i}, z_{j}) = \frac{z_{i}^{T}z_{j}}{\left\|z_{i}\right\| \cdot \left\|z_{j}\right\|}$. According to the clustering assumption \citep{assumption}, samples that are close to each other in the feature space should have the same predictions, so samples with locally inconsistent predictions are near decision boundaries and have a higher probability of falling into wrong clusters (dashed circle in Fig. \ref{fig2}). We calculate the local inconsistency degree $\mathcal{C}_{i}$ by counting the number of neighbors that have different pseudo labels from the query:
\begin{equation}
    \mathcal{C}_{i} = \sum_{j=1}^{k}|\hat{y}_{i} \neq \hat{y}_{\mathcal{N}_{i}^{j}}|
\end{equation}
where $\mathcal{N}_{i}^{j}$ is the index of the $j$-th neighbor of $x_{i}$.

To further select uncertain samples that are far away from cluster centers and near decision boundaries, we also restrict that the selected samples should have high prediction entropy. Specifically, we model the probability that samples belong to different clusters with Student’s t-distribution \citep{dec}, which can reflect the probability of the sample belonging to different clusters:
\begin{equation}
    q_{ij}=\frac{(1+\|z_i-\mu_j\|^2/\alpha)^{-\frac{\alpha+1}2}}{\sum_{j'}(1+\|z_i-\mu_{j'}\|^2/\alpha)^{-\frac{\alpha+1}2}}
\end{equation}
where $\alpha$ is the degree of freedom. If the distribution is uniform, the probability that the sample belongs to a specific cluster will be low and the sample can be far away from cluster centers and near decision boundaries, which can be easily misclassified. So we use the entropy of t-distribution to measure the degree of uniformity and select candidate labeling samples that have high entropy. The entropy can be calculated as:
\begin{equation}
    \mathcal{H}_{i}= - \sum_{j}q_{ij}log(q_{ij})
\end{equation}
Then we select a set of informative samples that have both high local inconsistency degree and prediction entropy:
\begin{equation}
    S = \{z_{i} \mid \mathcal{C}_{i} \in top_{m}(\mathcal{C}) \land \mathcal{H}_{i} \in top_{m}(\mathcal{H}) \}
\end{equation}
where $\mathcal{C}=\{\mathcal{C}_{j}\}_{j=1}^{N}$ and $\mathcal{H}=\{\mathcal{H}_{j}\}_{j=1}^{N}$ are the set of local inconsistency degree and the set of prediction entropy for each sample, respectively. $m$ is a hyperparameter that determines the number of samples to be selected.

\noindent \textbf{Discussion.}\quad
The proposed \textit{LIS} is effective in two aspects. First, the local inconsistency degree can help to select samples whose neighbors have the most diverse cluster assignments. Since these neighbor-chaotic samples may be located near decision boundaries and violate the clustering assumption \citep{assumption}, it will be hard for the model to decide which clusters they truly belong to. Second, the prediction entropy can select samples that are distributed uniformly among several clusters. Since these samples are far away from cluster centers and distributed near decision boundaries, they can be easily assigned to wrong clusters. By combing the two scores together, our model can select samples that are assigned to wrong clusters, and correcting these samples can provide more gains in improving model performance (Sec. \ref{lis}).

\subsection{Scalable Query Strategy}
\label{query}
Given the selected samples, the next step is how to query LLMs to get proper supervision information. However, we cannot directly query LLMs for categories because there is no label information for novel categories and the returned categories are hard to be aligned with the cluster assignments. So inspired by recent work \citep{clusterllm}, we propose a \textit{Scalable Query} strategy to mitigate the local inconsistent issue by querying LLMs which samples are the true neighbors of the selected samples. In this way, we can find the true cluster assignments of the selected samples by determining the neighborhood relationship between samples. This query strategy is scalable since we can set a different number of neighbor options for LLMs to choose from. Taking the query with $|q|$ options as an example, the prompt can be designed as: ``Select the sentence that better corresponds with the query sentence. Query: [$S$]. Sentence 1: [$S_{1}$]; Sentence 2: [$S_{2}$]; ...; Sentence $|q|$: [$S_{|q|}$].'', where [$S$] is the selected query sample and [$S_{1}$], [$S_{2}$] ... [$S_{|q|}$] are neighbor sentences of [$S$] from the top $|q|$ clusters that have the most neighbors of the query sample.

\noindent \textbf{Discussion.}\quad
The proposed query strategy can help to correct the local inconsistent samples by selecting their true neighbors from the chaotic neighborhood. This strategy is scalable since we can add a different number of options to query LLMs. Although adding more options will provide a higher probability of selecting the sample that is from the same category as the query, it will increase the query cost by adding more query tokens (Sec. \ref{sq}). Even if we do not find the true neighbor samples, our model can still learn semantic knowledge by pulling similar samples closer.

\subsection{Refined Neighborhood Contrastive Learning}
\label{rncl}
Based on the feedback of LLMs, we can refine the neighborhood relationships between samples. For the unselected samples, we randomly select a sample from their neighbors. In this way, these unselected samples can learn from diverse neighbors at different epochs rather than learn from a fixed neighbor, and the diversity of selected neighbors can enhance the generalization of our model. Then we can correct the selected samples and learn clustering-friendly representations by pulling samples closer to their neighbor samples with neighborhood contrastive learning \citep{ncl}:

\begin{equation}
\begin{split}
    \mathcal{L} &= -\frac{1}{N}\sum_{i=1}^{N}  log \frac{exp(\mathcal{A}^{T}(z_{i})\mathcal{A}(z_{\mathcal{N}_{i}^{s}})/\tau)}{\sum\limits_{z_{j} \in \mathcal{B}}exp(\mathcal{A}^T(z_{i})\mathcal{A}(z_{j})/\tau)} \\
\end{split}
\end{equation}
where $\mathcal{A}$ is a data augmentation method, $\mathcal{N}_{i}^{s}$ is the index of the selected neighbor of $z_{i}$, $\tau$ is a hyperparameter and $\mathcal{B}$ is the current batch. We also add cross-entropy loss $\mathcal{L}_{ce}(\mathcal{D}^{l})$ for training to enhance our model performance on known categories.

\subsection{Cluster Interpretation}
\label{interpret}
Different from previous works that only performed clustering to discover clusters without any semantic information, we propose to interpret the discovered clusters with the help of LLMs. Specifically, we first utilize the \textit{`Alignment and Decoupling'} strategy \citep{dpn} to decouple clusters that correspond to novel categories from the discovered clusters. 
Then for each decoupled cluster, we select a few samples that are closest to the center of the clusters as representative samples. Next, we make LLMs to summarize these samples to generate label names for these novel categories. Experimental results show that this strategy can select representative samples and generate accurate label names for the discovered novel categories (Sec. \ref{name}).

\subsection{Resource Saving}
By selecting the most informative samples and reducing the query options, our framework can reduce query cost. To further reduce the computing and query cost, we propose two strategies for our model training.

\noindent \textbf{Interval Update.}\quad Since the neighborhood relationships between samples will not change dramatically, we query LLMs and update the neighborhood relationships every few epochs (5 in our experiments). In this way, we can save the computing resources of neighborhood retrieval and the cost of querying LLMs.

\noindent \textbf{Query Result Storage.}\quad Since we may query LLMs for the same sample repeatedly in different epochs, we maintain a dictionary to store the query results to avoid duplicated queries. In this way, we can reuse the query results and reduce the cost of queries.

\begin{table*}
\setlength\tabcolsep{5.5pt}
\centering
\begin{tabular}{lccccccccc}
\toprule
\multirow{2}*{Method} & \multicolumn{3}{c}{BANKING} &\multicolumn{3}{c}{StackOverflow} & \multicolumn{3}{c}{CLINC}\\ 
\cmidrule(r){2-4}  \cmidrule(r){5-7}  \cmidrule(r){8-10}
            &H-score    &Known    &Novel    &H-score    &Known    &Novel    &H-score    &Known    &Novel \\
\midrule
DeepCluster &13.97  &13.94  &13.99  &19.10  &18.22  &14.80  &26.48  &27.34  &25.67  \\
DCN         &16.33  &18.94  &14.35  &29.22  &28.94  &29.51  &29.20  &30.00  &28.45  \\
DEC         &17.82  &20.36  &15.84  &25.99  &26.20  &25.78  &19.78  &20.18  &19.40  \\
KM-BERT     &21.08  &21.48  &20.70  &16.93  &16.67  &17.20  &34.05  &34.98  &33.16  \\
AG-GloVe    &30.47  &29.69  &31.29  &29.95  &28.49  &31.56  &44.16  &45.17  &43.20  \\
SAE         &37.77  &38.29  &37.27  &62.65  &57.36  &69.02  &45.74  &47.35  &44.24  \\

\midrule
Simple   &40.52  &49.96  &34.08  &57.53  &57.87  &57.20  &62.76  &70.60  &56.49  \\
Semi-DC  &47.40  &53.37  &42.63  &64.90  &63.57  &61.20  &73.41  &75.60  &71.34  \\
Self-Label    &48.19  &61.64  &39.56  &59.99  &78.53  &48.53  &61.29  &80.06  &49.65  \\
CDAC+    &50.28  &55.42  &46.01  &75.78  &77.51  &74.13  &69.42  &70.08  &68.77  \\
DTC      &52.13  &59.98  &46.10  &63.22  &80.93  &51.87  &68.71  &82.34  &58.95  \\
Semi-KM  &54.83  &73.62  &43.68  &61.43  &81.02  &49.47  &70.98  &89.03  &59.01  \\
DAC      &54.98  &69.60  &45.44  &63.64  &76.13  &54.67  &78.77  &89.10  &70.59  \\
GCD      &55.78  &75.16  &44.34  &64.63  &82.00  &53.33  &63.08  &89.64  &48.66  \\
PTJN     &60.69  &77.20  &50.00  &77.48  &72.80  &82.80  &83.34  &91.79  &76.32  \\
DPN      &60.73  &80.93  &48.60  &83.13  &85.29  &81.07  &84.56  &92.97  &77.54  \\
MTP      &61.59  &80.08  &50.04  &77.23  &84.75  &70.93  &80.32  &91.69  &71.46  \\
\midrule
\textbf{\textit{Loop} (Ours)}        &\textbf{74.60}  &\textbf{83.99}  &\textbf{67.10}  &\textbf{91.57}  &\textbf{87.56}  &\textbf{90.53}  &\textbf{90.74}  &\textbf{94.45}  &\textbf{87.31}  \\
\rowcolor{gray!20}Improvement    &+13.01    &+3.06  &+17.06  &+8.44  &+2.27  &+7.73  &+6.18  &+1.48  &+9.77  \\
\bottomrule

\end{tabular}
\caption{Model comparison results (\%) on the testing sets. Average results over 3 runs are reported. Some results are cited from \citet{dpn}.}
\label{table1}
\end{table*}

\section{Experiments}
\subsection{Experimental Setup}
\subsubsection{Datasets}
We perform experiments on three benchmark datasets.
\textbf{BANKING} \citep{banking} is an intent detection dataset in the bank domain.
\textbf{StackOverflow} \citet{stack} is a question classification dataset.  
\textbf{CLINC} \citep{clinc} is an intent detection dataset from multiple domains. For each dataset, we randomly select 25\% categories as novel categories and 10\% data as labeled data. More details are listed in Appendix \ref{data}.

\subsubsection{Comparison with SOTA Methods}
We compare our model with various baselines and SOTA methods.

\noindent \textbf{Unsupervised Models.}\quad  (1) DeepCluster \citep{deepcluster}. (2) DCN \citep{dcn}. (3) DEC \citep{dec}. (4) KM-BERT \citep{bert}. (5) AG-GloVe \citep{ag}. (6) SAE \citep{stackae}.

\noindent \textbf{Semi-supervised Models.}\quad (1) Simple \citep{simple}. (2) Semi-DC \citep{deepcluster}. (3) Self-Labeling: \citep{selflabel}. (4) CDAC+ \citep{thu2020}. (5) DTC \citep{dtc}. (6) Semi-KM \citep{bert} (7) DAC \citep{thu2021}. (8) GCD \citep{gcd}. (9) PTJN \citep{ptjn}. (10) DPN \citep{dpn}. (11) MTP \citep{mtp}.

\subsubsection{Evaluation Metrics}
We measure model performance with clustering accuracy with Hungarian algorithm \citep{hungarian} to align the predicted cluster labels and ground-truth labels.
(1) \textbf{H-score}: harmonic mean of the accuracy of known and novel categories \citep{hscore}.
(2) \textbf{Known}: accuracy of known categories.
(3) \textbf{Novel}: accuracy of novel categories.

\subsubsection{Implementation Details}
We use the pre-trained bert-base-uncased model \citep{huggingface} as the base model and the GPT-3.5 Turbo API as the LLM.
For hyper-parameters, $k$ is set to \{50, 50, 500\} for BANKING, CLINC and StackOverflow, respectively. $\alpha$ is set to 1, $m$ is set to 500, $|q|$ is set to 2 and $\tau$ is set to 0.07. 
The pre-training epoch is set to 100 and the training epoch is set to 50. All experiments are conducted on a single NVIDIA 3090 GPU. The learning rate for pretraining and training is set to $5e^{-5}$ and $1e^{-5}$, respectively.
For masked language modeling, the mask probability is set to 0.15 and Random Token Replace is used for data augmentation, following previous works \citep{mtp}. The used prompts in the paper are listed in Appendix \ref{prompt}.

\begin{table}
\centering
\begin{tabular}{lccc}
\toprule
Model & H-score & Known & Novel \\
\midrule  
\rowcolor{gray!20} \textit{Loop} (Ours) & 74.60 & 83.99 & 67.10\\
\midrule
ClusterLLM & 71.96 & 78.66 & 66.32\\
\midrule
w/o $\mathcal{L}_{ce}$  & 72.77 & 82.43 & 65.13 \\
w/o LLMs                & 70.02 & 78.15 & 63.42 \\
\midrule
\midrule
w/ Entropy              & 74.06 & 84.07 & 66.18 \\
w/ Margin			      & 72.88 & 82.73 & 65.13 \\
w/ Random			      & 72.33 & 82.23 & 64.56 \\
w/ Confidence			  & 72.08 & 82.44 & 64.03 \\
\midrule
\midrule
$|q| = 3$               & 75.91 & 84.25 & 69.08 \\
$|q| = 4$               & 77.30 & 83.84 & 71.71 \\
\midrule
OverClustering            & 74.07 & 80.54 & 68.56 \\
\bottomrule

\end{tabular}
\caption{Ablation study with different model variants.}
\label{table2}
\end{table}

\section{Experimental Analysis}
\subsection{Main Results}
\label{main}
We show the comparison results in Table \ref{table1}. From the results, we can see that our model gets the best performance on all datasets and evaluation metrics (average \textbf{7.67\%} improvement), which can show the effectiveness of our model. Specifically, our model gains average \textbf{9.21\%} improvement in H-score, which means that our model can better balance model performance on known and novel categories and alleviate the effects of model bias towards known categories. Average \textbf{2.27\%} improvement in the accuracy of known categories shows that our model can acquire semantic knowledge from both labeled and unlabeled data to enhance our model performance. Last but not least, our model gains average \textbf{11.52\%} improvement in the accuracy of novel categories. We attribute the remarkable improvement to the following reasons. First, \textit{Local Inconsistent Sampling} can help to select samples that have a higher probability of falling to the wrong clusters. Correcting them can provide more information gain for the model training. Second, \textit{Scalable Query} can provide accurate supervision by choosing the true neighbors, which can help mitigate the local inconsistency problem. Last, \textit{Neighborhood Contrastive Learning} with the refined neighbors can help to pull samples from the same category closer and learn clustering-friendly representations.

\subsection{Ablation Study}
We validate the effectiveness of components in our model on the BANKING dataset in Table \ref{table2}.
\subsubsection{Main Components}
From the results we can see that our model outperforms ClusterLLM \citep{clusterllm} since we can select more informative samples and provide more difficult candidate samples to query LLMs. 
Removing cross-entropy loss $\mathcal{L}_{ce}$ can lead to slight performance degradation since it is responsible for providing accurate supervision for known categories. And removing feedback from LLMs will lead to severe performance decline on both known and novel categories, which can reflect the importance of introducing LLMs to the training loop to provide supervision information.

\begin{table}
\centering
\begin{tabular}{lccc}
\toprule
Strategy & BANK. & Stack. & CLINC \\
\midrule  
Random			      & 33.00 & 20.50 & 17.50 \\
Margin			      & 77.00 & 68.50 & 68.50 \\
Entropy               & 80.00 & 84.50 & 61.00 \\
Confidence			  & 81.00 & 78.00 & 66.50 \\
\midrule
\textbf{\textit{LIS} (Ours)}  & \textbf{88.48} & \textbf{90.97} & \textbf{72.25} \\
\rowcolor{gray!20}    Improvement           & +7.48 & +6.47 & +3.75 \\
\bottomrule

\end{tabular}
\caption{Proportion of the selected 200 samples that fall into wrong clusters.}
\label{table3}
\end{table}

\subsubsection{Analysis of \textit{LIS}}
\label{lis}
To validate the proposed \textit{Local Inconsistent Sampling (LIS)} strategy, we compare the model performance with different sampling strategies. As shown in Table \ref{table2}, \textit{LIS} outperforms other sampling strategies, which demonstrates the effectiveness of our \textit{LIS} strategy. To further validate the proposed \textit{LIS} strategy, we also compare the accuracy of different strategies for selecting samples that fall into wrong clusters. From Table \ref{table3} we can see that \textit{LIS} outperforms other strategies by a large margin, which means that \textit{LIS} can select more informative samples to boost our model performance.

\begin{figure*} \centering    
\subfigure[Accuracy of the corrected samples.] {
 \label{fig:a}     
\includegraphics[width=0.66\columnwidth, height=4cm]{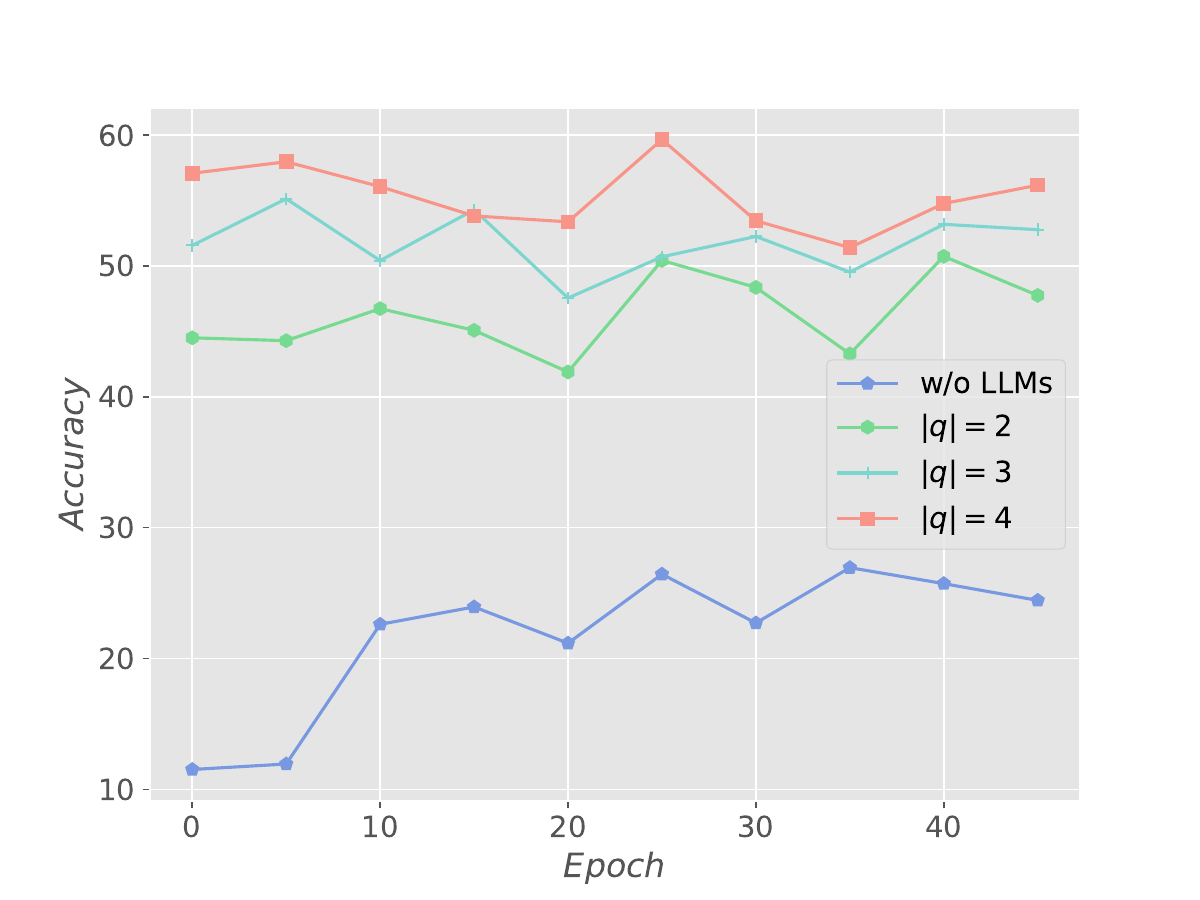}  
}     
\subfigure[Effect of the number of query samples.] { 
\label{fig:b}     
\includegraphics[width=0.66\columnwidth, height=3.5cm]{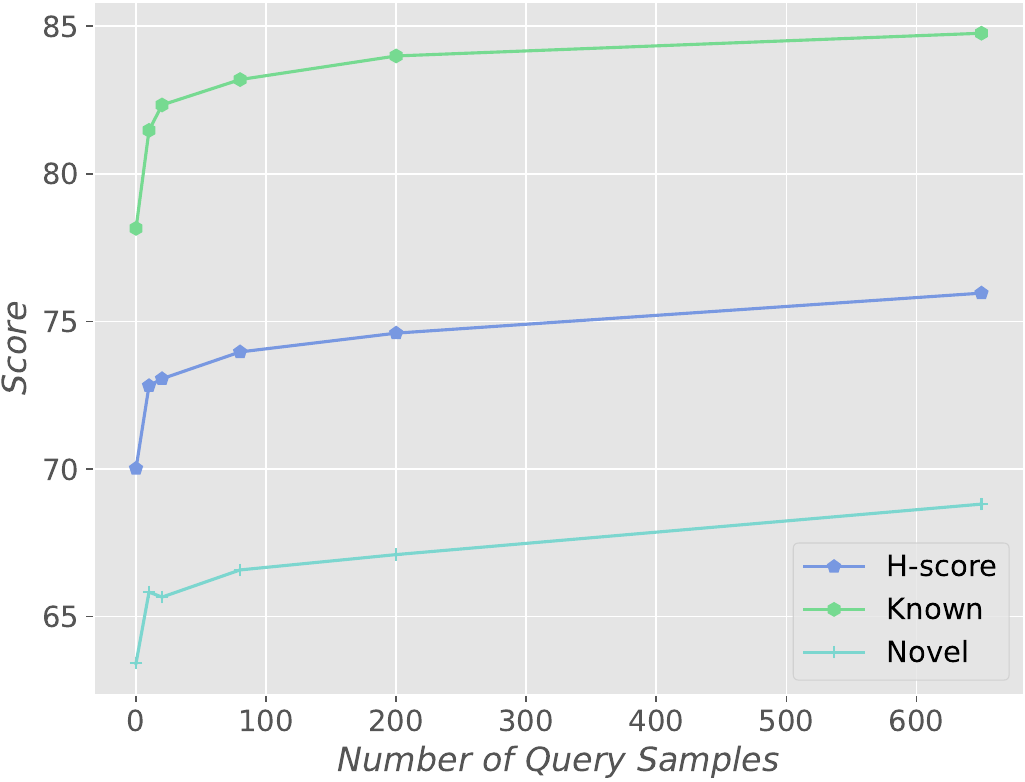}     
}    
\subfigure[t-SNE Visualization.] { 
\label{fig:c}     
\includegraphics[width=0.66\columnwidth, height=3.5cm]{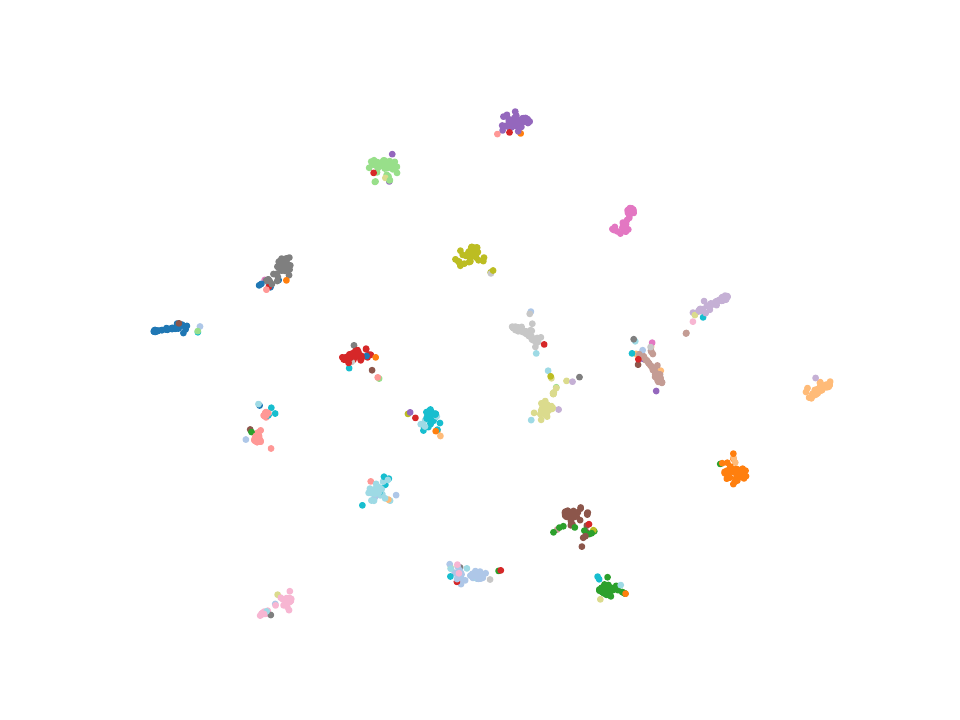}     
}
\caption{Analysis of the quality of representation learning and neighborhood retrieval.}     
\label{figure3}     
\end{figure*}

\begin{table*}[t]
\centering
\setlength\tabcolsep{15pt}
\begin{tabular}{lcc}
\hline
\multicolumn{1}{c}{Selected Sentences} & Ground Truth & Prediction \\
\midrule
Can I change my PIN if I want to?  &\multirow{3}*{Change PIN} &\multirow{3}*{Change PIN} \\  Can I change my PIN?  \\ Do I have to change my PIN at a bank? \\
\midrule
What will the weather be this weekend?  &\multirow{3}*{Weather} &\multirow{3}*{Weather forecast} \\ Tell me what the weather is like.  \\ What's the weather like? \\
\bottomrule
\end{tabular}
\caption{Examples of the selected sentences, ground-truth category names and predicted category names.}
\label{table5}
\end{table*}

\begin{table}
\centering
\begin{tabular}{cccc}
\toprule
$|q|$        & 2      & 3      & 4        \\
\midrule
Cost (\$)    & 0.39   & 0.47   & 0.55     \\
\bottomrule
\end{tabular}
\caption{Query cost with different number of options.}
\label{table4}
\end{table}

\subsubsection{Analysis of \textit{Scalable Query}}
\label{sq}
To validate the \textit{Scalable Query} strategy, we compare the model performance with the different number of options $|q|$. As shown in Table \ref{table2}, Increasing $|q|$ can improve our model performance because we can select more accurate neighbors as $|q|$ grows. As shown in Fig. \ref{fig:a}, our scalable query strategy can correct many samples compared to the method without LLM queries, which shows the effectiveness of our query strategy. With $|q|$ increasing, our model can correct more samples and get better model performance, which shows the scalability of our query method. However, the query cost will also increase with the growth of $|q|$ due to the growth of query tokens (Table \ref{table4}), so the \textit{Scalable Query} strategy provides users with flexible options to balance query cost and model performance.

\subsection{Cluster Interpretation}
\label{name}
In addition to the improved model performance, our model can also interpret the discovered clusters by generating category names for them. As shown in Table \ref{table5}, our model can select representative samples for novel categories and generate accurate names for them, which can provide more convenience for real-world applications of our model. More results are listed in Appendix \ref{inter}.

\subsection{Influence of the Number of Samples}
We investigate the influence of the number of selected samples for query in Fig. \ref{fig:b}. From the results, we can see that increasing the number of samples can improve our model performance. However, the growth rate gradually slows down because it becomes increasingly difficult to select informative samples as the number of samples increases.

\begin{table}[t]
\centering
\linespread{0.5}
\begin{tabular}{lccc}
\toprule
Method & BANK.  & Stack.  & CLINC \\
\midrule
Ground Truth       & 77    & 20 & 150 \\
\midrule
DAC                & 66    & 15 & 130 \\
DPN                & 67    & 18 & 137 \\
\midrule
\rowcolor{gray!20} \textbf{Ours}      & \textbf{78}    & \textbf{19} & \textbf{145} \\
\bottomrule
\end{tabular}
\caption{Estimation of the number of categories.}
\label{table6}
\end{table}

\subsection{Real-world Applications}
\label{estimate}
In the real world, the number of categories $K$ is usually unknown. To solve this issue, we utilize the filtering strategy \citep{thu2021} to estimate $K$. As shown in Table \ref{table6}, our model obtains the most accurate estimation with only a little error, which shows the effectiveness of our model. To further investigate the influence of $K$, we perform over-clustering by over-estimating $K$ used for inference by a factor of 1.2. Results in Table \ref{table2} show that our model gets close performance even without knowing the ground truth $K$, which validates the robustness of our model.

\subsection{Influence of the Number of Neighbors}
\begin{figure}[t]
\centering
\includegraphics[width=7.5cm]{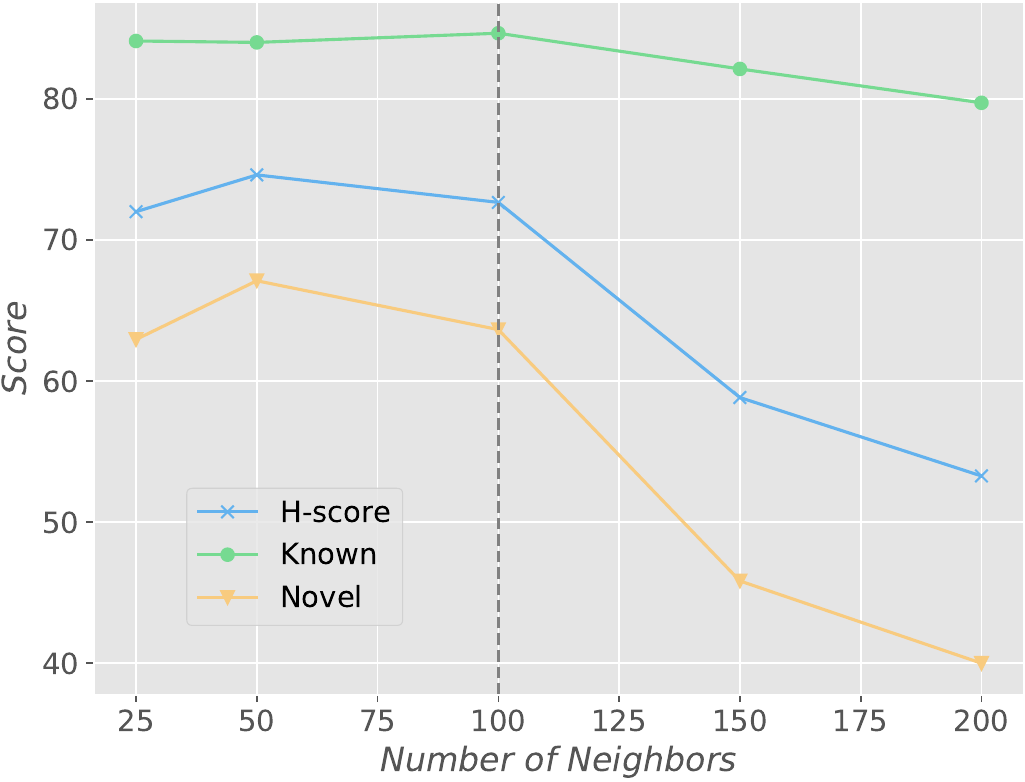}
\caption{Model performance with the different number of neighbors.} 
\label{fig5}
\end{figure}

To investigate the influence of the number of neighbors $k$, we perform experiments with $k$ in the set $\{25, 50, 100, 150, 200\}$ on the BANKING dataset. As shown in Fig. \ref{fig5}, our model gets a similar performance when $k$ is less than 100. However, when $k$ exceeds 100 by a lot, our model performance drops quickly. This is because when $k$ exceeds the average number of samples for each category by a lot (e.g., approximately 110 for the BANKING dataset), there is a higher probability for neighborhood contrastive learning to randomly select samples from other categories as the positive key, which can introduce much noise for model training and degrade the model performance. Experiments about the influence of the known category ratio are listed in Appendix \ref{ratio}.

\subsection{Visualization}
We visualize the learned embeddings of our model on the StackOverflow dataset with the t-SNE technique in Fig. \ref{fig:c}. From the figure, we can see that our model can learn more separable clusters and decision boundaries for different categories, which indicates that our model can learn more discriminative features for clustering. More visualization results are shown in Appendix \ref{visualization}.

\section{Conclusion}
In this paper, we propose \textit{Loop}, an active-learning framework that introduces LLMs to the training loop for Generalized Category Discovery, which can boost our model performance without any human effort. We further propose \textit{Local Inconsistent Sampling} to select informative samples and utilize \textit{Scalable Query} to correct these samples with the feedback of LLMs. By pulling samples closer to their refined neighbors, our model can learn clustering-friendly representations. Finally, we generate label names for the discovered clusters to facilitate real-world applications. Experiments show that \textit{Loop} outperforms SOTA models by a large margin and generates accurate category names for the discovered clusters.

\section*{Limitations}
Even though the proposed \textit{Loop} framework achieves superior performance on the GCD task, it still faces the following limitations. First, when increasing the number of samples to query LLMs, the performance of \textit{Loop} improves slowly, which is because it becomes harder to select informative samples. So how to revise the sample selection strategy to select more informative samples is a key question. Second, the \textit{Scalable Query} can only provide neighborhood information, which is relatively weak supervision compared to category supervision in traditional active learning. So how to design a query strategy to acquire more accurate supervision is another key question. Last, \textit{Loop} relies on the feedback of LLM APIs, which is uncontrollable, and uploading data to query LLMs may be risky for some sensitive industries. So how to solve the above limitations will be our feature work.

\section*{Acknowledgments}
This work was supported by National Science and Technology Major Project (2022ZD0117102), National Natural Science Foundation of China (62293551, 62177038, 62277042, 62137002, 61937001,62377038). Project of China Knowledge Centre for Engineering Science and Technology, ‘‘LENOVO-XJTU’’ Intelligent Industry Joint Laboratory Project.

\bibliography{anthology}

\appendix
\section{Appendix}
\label{sec:appendix}
\subsection{Datasets}
\label{data}
To validate the effectiveness of our \textit{Loop} framework, we perform experiments on three benchmark datasets: \textbf{BANKING} \citep{banking}, \textbf{StackOverflow} \citep{stack} and \textbf{CLINC} \citep{clinc}.
For each dataset, we randomly select 25\% categories as novel categories and then select 10\% data from each known category as labeled data. 
After training, we test model performance on the testing set in an inductive manner. We also perform experiments with different known category ratios in Appendix \ref{ratio}. Statistics of the datasets are listed in Table \ref{table7}.

\begin{figure*}
\centering
\includegraphics[width=14cm]{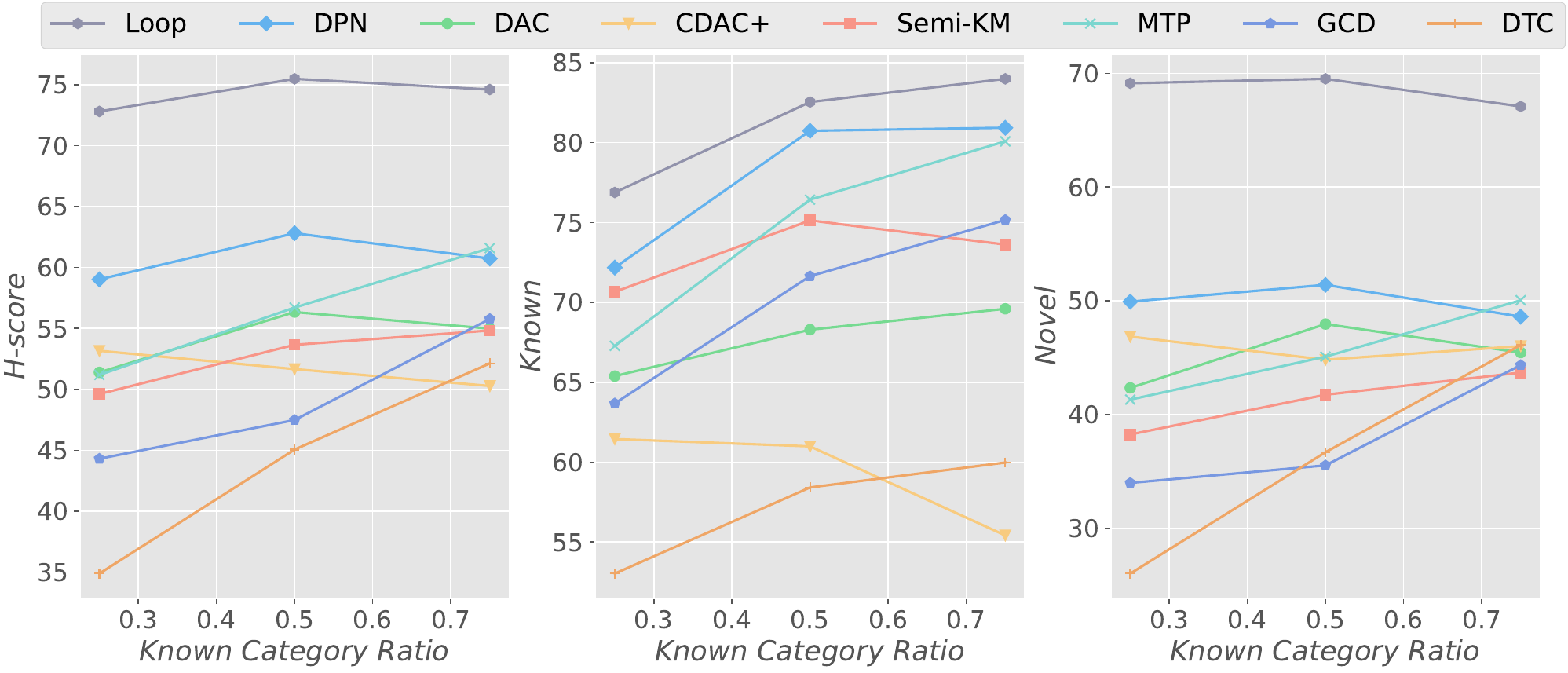}
\caption{Model performance with different known category ratios.} 
\label{fig4}
\end{figure*}

\subsection{Prompt Design}
\label{prompt}
\noindent \textbf{Query Prompt.} \quad Following \citet{clusterllm}, we design the query prompt as follows:

``Select the sentence that better corresponds with the query sentence in terms of intents or categories. Please respond with 'Sentence 1' or 'Sentence 2' ... or 'Sentence $|q|$' without explanation. 

Query: [$S$]. Sentence 1: [$S_{1}$]; Sentence 2: [$S_{2}$]; ...; Sentence $|q|$: [$S_{|q|}$].''

\noindent \textbf{Interpretation Prompt.} \quad To generate category names for the discovered clusters that correspond to novel categories, we select three samples that are closest to the center of the clusters as representative samples. And we design the interpretation prompt as follows:

``Given the following sentences, return a word or a phrase to summarize the common intent or category of these sentences without explanation. 

Sentence 1: [$S_{1}$]; Sentence 2: [$S_{2}$]; Sentence 3: [$S_{3}$].''

\begin{table}[t]
\centering
\begin{tabular}{lccccc}
\hline
Dataset & $|\mathcal{Y}_{k}|$ & $|\mathcal{Y}_{n}|$ & $|\mathcal{D}^{l}|$ & $|\mathcal{D}^{u}|$ & $|\mathcal{D}^{t}|$\\
\hline
BANK.         &  58    &  19     & 673    & 8,330    & 3,080\\
Stack.        &  15    &  5      & 1,350  & 16,650   & 1,000\\
CLINC         &  113   &  37     & 1,344  & 16,656   & 2,250\\
\hline
\end{tabular}
\caption{Statistics of datasets. $|\mathcal{Y}_{k}|$, $|\mathcal{Y}_{n}|$, $|\mathcal{D}^{l}|$, $|\mathcal{D}^{u}|$ and $|\mathcal{D}^{t}|$ represent the number of known and novel categories, labeled, unlabeled and testing data, respectively.}
\label{table7}
\end{table}

\subsection{Influence of the Known Category Ratio}
\label{ratio}
In the real world, the ratio of known categories may vary in different applications and the number of novel categories may exceed the number of known ones. To validate the robustness of our model towards the changing known category ratios, we perform experiments with known category ratios in the set $\{25\%, 50\%, 75\%\}$ on the BANKING dataset. As shown in Fig. \ref{fig4}, our model gets the best performance on all known category ratios and evaluation metrics, which can show the effectiveness and robustness of our model towards different known category ratios. Furthermore, our model outperforms other methods by a large margin on the accuracy of novel categories and H-score, which can further validate that our \textit{Loop} framework can learn better representations based on the feedback of LLMs.

\begin{figure*}[t] 
\centering    
\subfigure[MTP.] {
 \label{fig:d}     
\includegraphics[width=0.66\columnwidth, height=4cm]{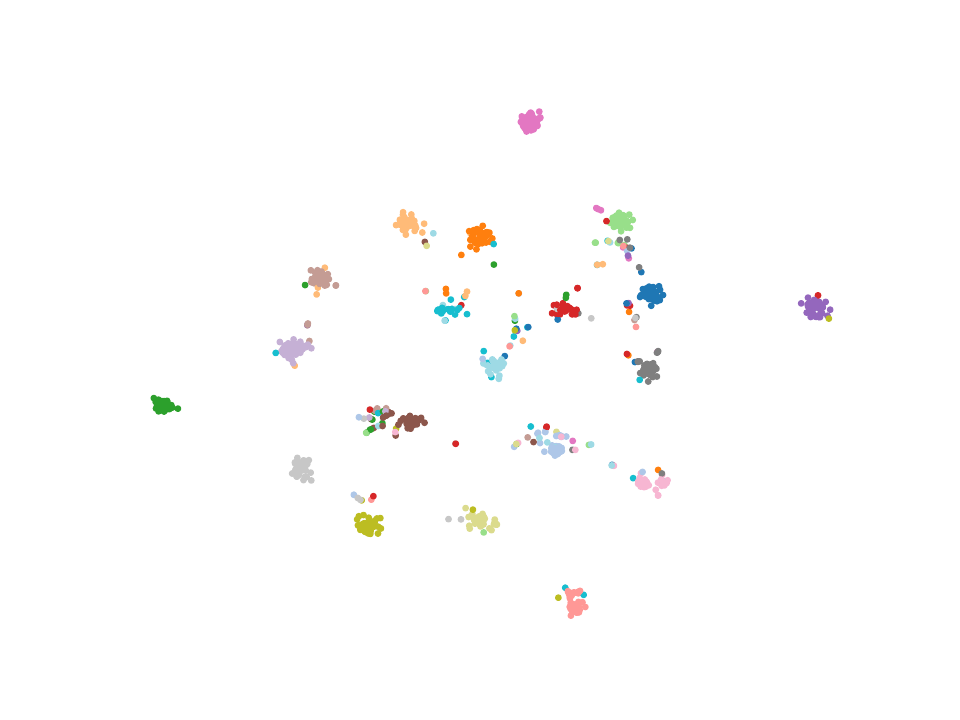}  
}     
\subfigure[DPN.] { 
\label{fig:e}     
\includegraphics[width=0.66\columnwidth, height=4.5cm]{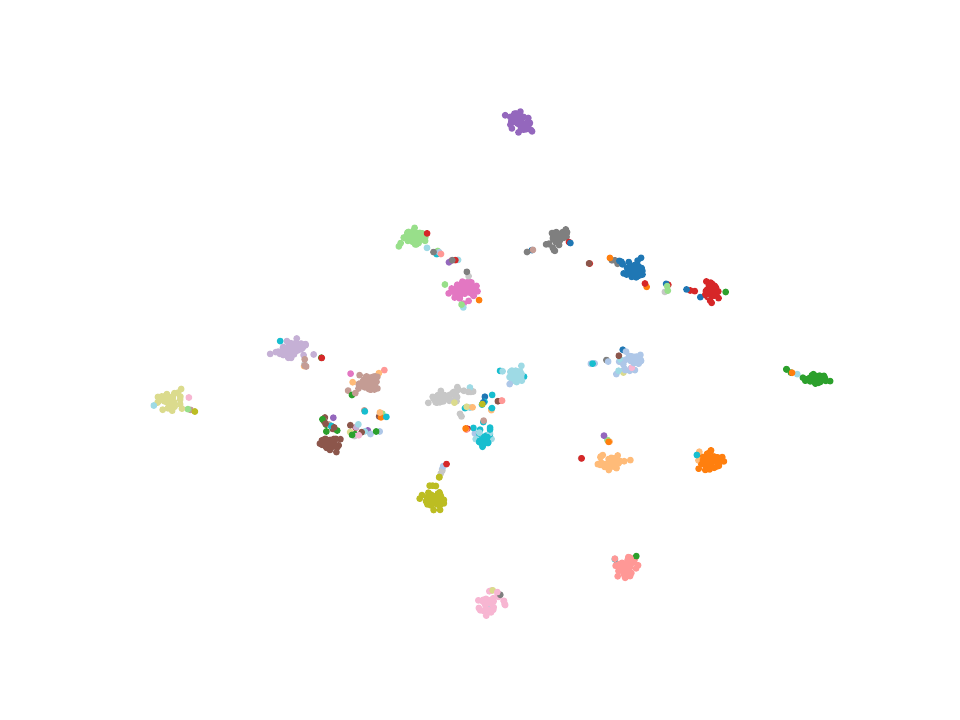}     
}    
\subfigure[\textit{Loop} w/o LLM query.] { 
\label{fig:f}     
\includegraphics[width=0.66\columnwidth, height=3.7cm]{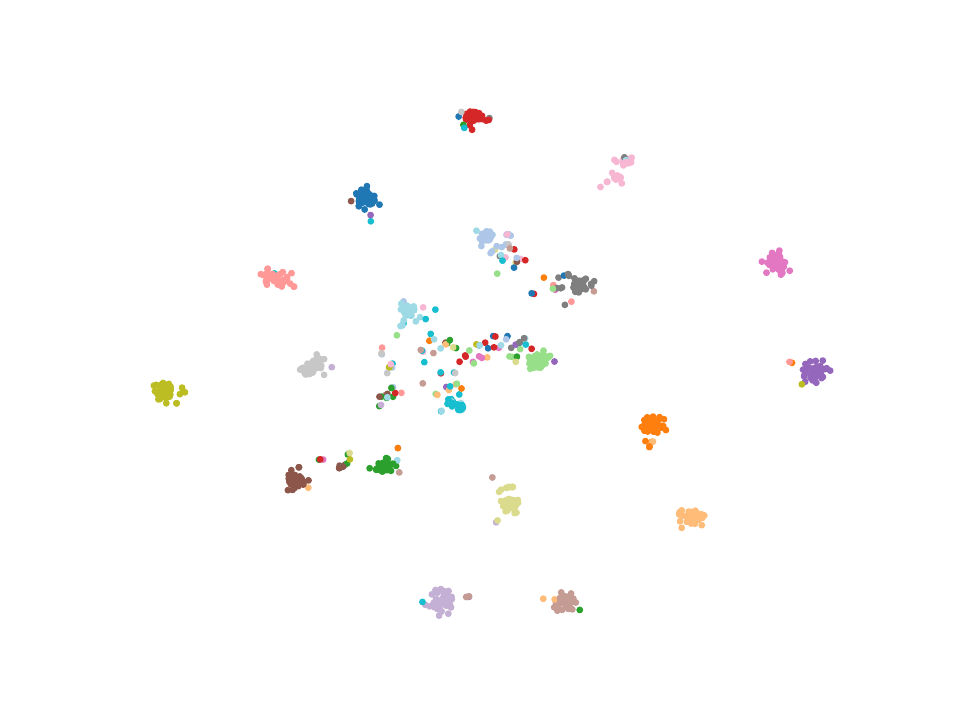}     
}
\caption{t-SNE Visualization for the compared methods.}     
\label{figure6}     
\end{figure*}

\subsection{More Results of Feature Visualization}
\label{visualization}
We also visualize the learned embeddings by previous SOTA methods (MTP and DPN) on the StackOverflow dataset in Fig. \ref{figure6}. Compared to the visualization results of our model in Fig. \ref{fig:c}, we can see that some clusters are mixed together for the compared methods, which can indicate that our model can learn more discriminative features and form more separatable decision boundaries for different categories. Furthermore, if we remove the feedback of LLMs from our model (\textit{Loop} w/o LLM query), clusters corresponding to novel categories will be mixed together due to the lack of supervision, which can further validate the effectiveness of our active-learning framework.

\subsection{More Results of Cluster Interpretation}
\label{inter}
We provide more examples of the selected sentences and generated category names on the three datasets in Table \ref{table8}. The results can further validate that our model can select representative samples and generate accurate names for the discovered novel categories, which can validate the effectiveness of our interpretation strategy.

\begin{table*}
\centering
\begin{tabular}{lcc}
\hline
\multicolumn{1}{c}{Selected Sentences} & Ground Truth & Prediction \\
\midrule
Can I top up with check?  &\multirow{3}*{\makecell[c]{Top up by cash \\ or check}} &\multirow{3}*{\makecell[c]{Top up with a check}} \\  Where do I find how to top off with a check?  \\ Can I top up my account with a check? \\
\midrule
Why is my cash withdrawal still showing as pending?  &\multirow{3}*{\makecell[c]{Pending cash \\ withdrawal}} &\multirow{3}*{\makecell[c]{Pending cash \\ withdrawal}} \\  My cash withdrawal is showing as pending, why?  \\ My cash withdrawal shows as pending still. \\

\midrule
\midrule
Look up the calories in an apple.  &\multirow{3}*{Calories} &\multirow{3}*{\makecell[c]{Calorie information \\ for food}} \\  What's the amount of calories in a cheesy omelette?  \\ Look up the calories in Cheetos. \\
\midrule
Tell me how much my state taxes amount to.  &\multirow{3}*{Taxes} &\multirow{3}*{Tax amount} \\ What is the amount of my state taxes?  \\ What is the amount of my federal taxes? \\
\midrule
\midrule
How to convert excel sheet column names?  &\multirow{3}*{Excel} &\multirow{3}*{\makecell[c]{Excel functionalities \\ and operations}} \\  Setup an excel template.  \\ How do you prevent printing dialog of excel? \\
\midrule
How to pass URL variables into a WordPress page?  &\multirow{3}*{WordPress} &\multirow{3}*{\makecell[c]{WordPress \\ Customization}} \\  Get three posts before a certain date in WordPress.  \\ Where to place a query to show posts in wordpress? \\
\bottomrule
\end{tabular}
\caption{Examples of the selected sentences, ground-truth category names and predicted category names.}
\label{table8}
\end{table*}


\end{document}